\documentclass[journal,twoside,web]{ieeecolor}
\usepackage{generic}
\usepackage{cite}
\usepackage{amsmath,amssymb,amsfonts}
\usepackage{algorithmic}
\usepackage{graphicx}
\usepackage{algorithm,algorithmic}
\usepackage[hidelinks]{hyperref}
\usepackage{textcomp}
\usepackage{booktabs}
\usepackage{multirow}
\usepackage{xcolor}
\usepackage{adjustbox}
\usepackage{dsfont}
\usepackage{bbm}

\newcommand{\improve}[1]{~{\footnotesize \color{wincolor} ($\uparrow$#1)}}

\definecolor{wincolor}{RGB}{0, 150, 0} 
\definecolor{losecolor}{RGB}{200, 0, 0} 

\def\BibTeX{{\rm B\kern-.05em{\sc i\kern-.025em b}\kern-.08em
    T\kern-.1667em\lower.7ex\hbox{E}\kern-.125emX}}
\begin{document}

\title{Empowering Locally Deployable Medical Agent via State Enhanced Logical Skills for FHIR-based Clinical Tasks}
\author{\centering
Wanrong~Yang, Zhengliang~Liu, Yuan~Li, Bingjie~Yan, Lingfang~Li, Mingguang~He,
Dominik~Wojtczak, Yalin~Zheng, and~Danli~Shi
\thanks{Corresponding authors: Yalin Zheng, and Danli Shi. W. Yang, L. Li, and D. Wojtczak are with the Department of Computer Science, University of Liverpool, Liverpool, UK. B. Yan, M. He, and D. Shi are with The Hong Kong Polytechnic University, Hong Kong, China. Y. Li is with Beijing Normal University, Beijing, China. Z. Liu is with the School of Computing, University of Georgia, Athens, GA, USA. Y. Zheng is with the Department of Eye and Vision, University of Liverpool, Liverpool, UK.}}

\maketitle

\begin{abstract}
While Large Language Models demonstrate immense potential as proactive Medical Agents, their real-world deployment is severely bottlenecked by data scarcity under privacy constraints: clinical interaction trajectories are highly sensitive and cannot be used for cloud-based training, while acquiring the annotated multi-turn data required for reinforcement learning is prohibitively difficult. Consequently, traditional approaches that embed environment-specific workflows into model weights are often impractical for locally deployable models facing unseen clinical settings. To overcome this, we propose State-Enhanced Logical-Skill Memory (SELSM), a training-free framework that distills simulated clinical trajectories into entity-agnostic operational rules within an abstract skill space. During inference, a Query-Anchored Two-Stage Retrieval mechanism dynamically fetches these entity-agnostic logical priors to guide the agent's step-by-step reasoning, effectively resolving the ``state polysemy'' problem. Evaluated on MedAgentBench---the only authoritative high-fidelity virtual EHR sandbox benchmarked with real clinical data---SELSM substantially elevates the zero-shot capabilities of locally deployable foundation models (30B--32B parameters). Notably, on the Qwen3-30B-A3B backbone, our framework completely eliminates task chain breakdowns to achieve a 100\% completion rate, boosting the overall success rate by an absolute 22.67\% and significantly outperforming existing memory-augmented baselines. This study demonstrates that equipping models with a dynamically updatable, state-enhanced cognitive scaffold is a privacy-preserving and computationally efficient pathway for local adaptation of AI agents to clinical information systems. While currently validated on FHIR-based EHR interactions as an initial step, the entity-agnostic design of SELSM provides a principled foundation toward broader clinical deployment.
\end{abstract}


\section{Introduction}
\label{sec:introduction}

Realizing the clinical potential of Large Language Models (LLMs) demands a shift from static question-answering tools \cite{goodell2025large} to proactive \textit{Medical Agents} capable of executing complex clinical workflows. LLMs have already demonstrated significant promise in healthcare \cite{zou2025rise,meng2024application}, spanning clinical question answering \cite{singhal2025toward}, medical image analysis, and diagnostic decision-making \cite{wang2025large, yu2025large}. Yet systems with genuine clinical value must go further---comprehending high-level instructions, planning multi-step procedures, and invoking tools within Electronic Health Record (EHR) systems \cite{khashan2025understanding} to place medication orders, retrieve patient records, and schedule referrals, thereby alleviating professional shortages and administrative burdens \cite{cyril2025almanac, kim2025bilingual}.

However, deploying such agents in practice is fundamentally bottlenecked by \textbf{data scarcity under privacy constraints}, which prevents the prevailing parameter-updating paradigm from scaling to local clinical environments. Current approaches assume that enhancing tool-use capabilities requires costly parameter updates, either through continuous pre-training on massive medical corpora \cite{kim2025fine, liu2025application} or large-scale post-training via reinforcement learning (RL) \cite{lai2025med, wang2025survey}. Yet this paradigm, which statically embeds operational rules into model weights \cite{rouzrokh2025current}, is impractical for real-world deployment: clinical interaction trajectories are highly sensitive \cite{ashfaq2025artificial} and cannot be transmitted to cloud-based commercial models for training, local deployment of ultra-large models imposes prohibitive computational requirements for most hospitals \cite{li2024large, mayer2020scalable}, and acquiring the high-quality annotated multi-turn trajectories required for RL is extremely difficult under stringent privacy regulations \cite{yu2021reinforcement, lin2025training}. Moreover, because such approaches embed environment-specific strategies directly into model parameters \cite{yu2021reinforcement}, they are inherently brittle when confronting the heterogeneity of real-world clinical IT infrastructure---where different hospitals operate with distinct Fast Healthcare Interoperability Resources (FHIR) endpoints, customized formularies, and localized operational protocols \cite{jiang2025medagentbench, culler2006urban}---causing models to hallucinate, produce formatting errors, or fail entirely during multi-step execution \cite{bigoulaeva2025inherent}. These compounding barriers underscore the need for training-free approaches that can generalize without prior exposure to target-environment data.

Evaluation on MedAgentBench \cite{jiang2025medagentbench}---a high-fidelity virtual EHR sandbox whose benchmark tasks are grounded in real clinical data---exposes a stark reality: without environment-specific training data, even state-of-the-art models are far from achieving clinical reliability. Rather than viewing this benchmark merely as a standard evaluation dataset, we frame it as a rigorous testbed for evaluating \textbf{zero-shot deployment capabilities}. For example, Claude 3.5 Sonnet achieves only a 69.67\% success rate, frequently failing due to strict formatting constraints. More concerningly, smaller-scale locally deployable foundation models (e.g., in the 7B--70B parameter range) perform substantially worse, often suffering complete task chain breakdowns due to their inability to recover from intermediate API errors.

To address this zero-shot execution challenge, we propose an alternative to parameter updating: \textbf{equipping locally deployable models with a dynamically growing external memory of reusable operational logic}. Our key observation is that while surface-level states (e.g., patient demographics, API identifiers) vary across tasks and clinical environments, the underlying \textit{clinical operational logic} (e.g., verifying patient identity $\rightarrow$ checking allergy history $\rightarrow$ querying the formulary $\rightarrow$ submitting a prescription) is structurally consistent and finite. Based on this insight, we propose \textbf{SELSM}, a training-free framework that projects raw interaction trajectories into a compressed abstract skill space, distilling entity-agnostic operational rules as reusable ``Logical Skills.'' At inference time, a Query-Anchored Two-Stage Retrieval mechanism dynamically fetches relevant logical priors to guide step-by-step reasoning, without modifying model weights. Because the distilled skills are entity-agnostic by design, this framework also holds promise for cross-institutional deployment scenarios where each hospital generates institution-specific skills from local simulators, though we leave multi-site validation for future work.

The main contributions of this paper are summarized as follows: \textbf{A Training-Free Memory-Augmented Framework for Zero-Shot EHR Task Execution.} We propose SELSM, a framework that circumvents healthcare data scarcity by distilling simulated interaction trajectories into entity-agnostic logical skills, significantly improving the zero-shot EHR task execution of locally deployable LLMs (30B--32B parameters) without any parameter updates. \textbf{Query-Anchored Two-Stage Retrieval for State Polysemy Resolution.} We design a hierarchical retrieval mechanism grounded in entity-agnostic abstraction that resolves the ``state polysemy'' problem---where identical intermediate states arise across distinct clinical workflows---enabling precise skill matching at both task and transition levels. \textbf{Empirical Validation on a High-Fidelity EHR Sandbox.} Through evaluations on MedAgentBench, we demonstrate that SELSM markedly improves task completion and success rates, with the Qwen3-30B-A3B backbone achieving 100\% completion and a 22.67\% absolute gain in success rate over the vanilla baseline, significantly outperforming existing memory-augmented methods.

\section{Methods}
\label{sec:methods}

\begin{figure*}[!t]
\centerline{\includegraphics[width=\textwidth]{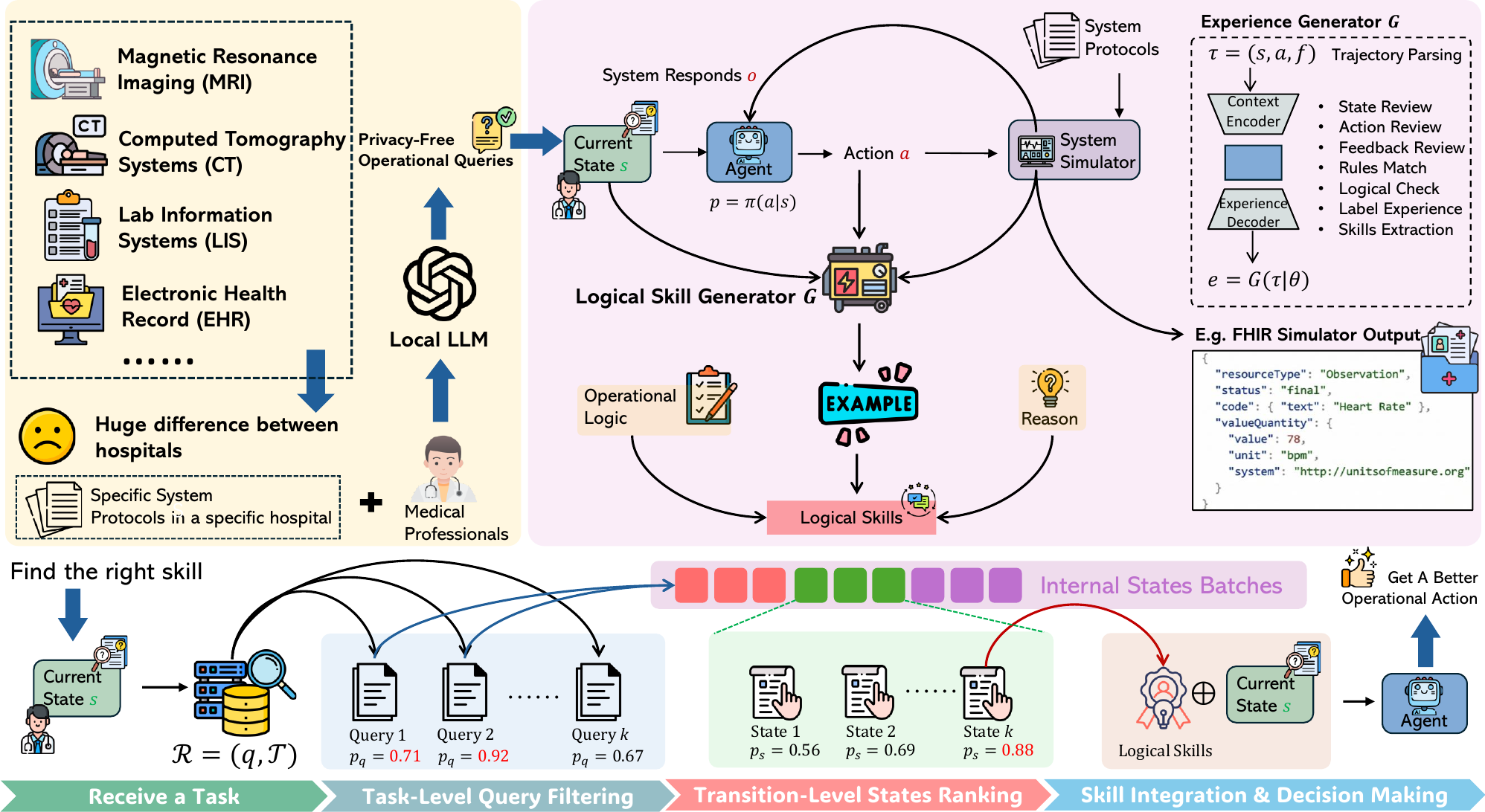}}
\caption{Overview of the SELSM framework. \textbf{(Top-left)}~Cross-institutional deployment context: heterogeneous hospital systems (EHR, LIS, PACS, etc.) operate under institution-specific protocols, while a locally deployed LLM, guided by medical professionals, issues privacy-preserving operational queries. \textbf{(Top-right)}~\textit{Logical Skill Distillation} (Phase~1): the agent interacts with a system simulator in a closed loop, where it observes the current state~$s$, executes an action~$a$ via the policy $p = \pi(a \mid s)$, and receives the system response~$o$. The Logical Skill Generator~$\mathcal{G}$ parses each trajectory $\tau = (s, a, o)$ through a context encoder and experience decoder to produce entity-agnostic logical skills $e = \mathcal{G}(\tau \mid \theta)$, comprising Operational Logic, canonical examples, and reasoning traces. \textbf{(Bottom)}~\textit{Query-Anchored Two-Stage Retrieval} (Phase~3): upon receiving a new task, the system first performs \textit{Task-Level Query Filtering} by scoring stored records $\mathcal{R} = (q, \mathcal{T})$ via query similarity~$p_q$, then executes \textit{Transition-Level State Ranking} over candidate internal states via state similarity~$p_s$, and finally integrates the retrieved skill with the current state to guide the agent toward a better operational action.}
\label{fig:framwork}
\end{figure*}

To realize the Logical Skills paradigm motivated in Section~\ref{sec:introduction}, we propose SELSM (\textbf{S}tate-\textbf{E}nhanced \textbf{L}ogical-\textbf{S}kill \textbf{M}emory), a training-free, memory-augmented framework that enables locally deployable LLM agents to acquire reusable clinical operational logic from simulated clinical system interactions (e.g., EHR environments) and generalize it to unseen medical settings at inference time. As illustrated in Fig.~\ref{fig:framwork}, SELSM operates in three phases: (1)~\textit{Logical Skill Distillation} (Fig.~\ref{fig:framwork}, top-right), which collects agent trajectories and abstracts them into entity-agnostic logical skills via a Logical Skill Generator~$\mathcal{G}$; (2)~\textit{Hierarchical Memory Indexing}, which organizes distilled skills into a two-level embedding index; and (3)~\textit{Query-Anchored Two-Stage Retrieval and Injection} (Fig.~\ref{fig:framwork}, bottom), which dynamically retrieves the most relevant skills to guide the agent's reasoning at each decision step, all without modifying the underlying model weights.

\subsection{Problem Formulation}
\label{sec:formulation}

We consider an LLM agent that interacts with an EHR system through FHIR APIs, as depicted in the agent--simulator loop of Fig.~\ref{fig:framwork} (top-right). At each time step~$t$, the agent observes a state~$s_t$ composed of the task instruction, clinical context, and the history of prior actions and system responses. The agent then produces an action~$a_t$, which is a FHIR API call (\texttt{GET}/\texttt{POST}) or a \texttt{FINISH} command, and the EHR system returns a response~$o_t$. The interaction continues for up to~$T$ rounds or until the agent issues \texttt{FINISH}.

Formally, a task episode is defined as a trajectory
\begin{equation}
    \tau = \bigl(q,\; c,\; \{(s_t, a_t, o_t)\}_{t=1}^{T}\bigr),
    \label{eq:trajectory}
\end{equation}
where $q$ denotes the clinical query (instruction) and $c$ denotes the patient context. The objective of SELSM is to construct a reusable logical skill memory~$\mathcal{M}$ from training trajectories and leverage it to improve the agent's decision-making on unseen tasks at inference time.

\subsection{Phase 1: Logical Skill Distillation}
\label{sec:phase1}

The first phase transforms raw agent--environment interaction traces into abstract, entity-agnostic logical skills, corresponding to the upper-right region of Fig.~\ref{fig:framwork}. It consists of three stages: trajectory collection, step-level evaluation, and memory record construction.

\subsubsection{Agent--Environment Interaction Loop}
During the training phase, the agent solves clinical tasks by interacting with a simulated FHIR server. For each training task~$(q, c)$, we maintain a cumulative state representation~$s_t$ that tracks the full interaction history. The initial state is constructed as
\begin{equation}
    s_0 = \mathcal{I}(q,\, c),
    \label{eq:s0}
\end{equation}
where $\mathcal{I}(\cdot)$ denotes the initialization function that composes the task instruction~$q$ and patient context~$c$ into the agent's initial observation. At each subsequent round, the state is updated via
\begin{equation}
    s_t = s_{t-1} \oplus \mathcal{F}(t,\, a_t,\, o_t), \quad t = 1, \ldots, T,
    \label{eq:st}
\end{equation}
where $\oplus$ denotes string concatenation and $\mathcal{F}(\cdot)$ is a formatting function that serializes the round index~$t$, the agent's action~$a_t$, and the FHIR system response~$o_t$ into a structured textual record. The episode terminates when $a_t$ contains the keyword \texttt{FINISH}. To increase the diversity of collected trajectories, each training task is executed~$K$ times (typically $K{=}3$) with temperature-based sampling, producing multiple distinct trajectories per task.

\subsubsection{LLM-as-Judge Step-Level Evaluation}
\label{sec:judge}
An innovation of SELSM is the use of an LLM-as-Judge~$\mathcal{J}$ to evaluate \emph{every step} of the agent's trajectory, rather than only the final outcome. This judge constitutes the core of the Logical Skill Generator~$\mathcal{G}$ depicted in Fig.~\ref{fig:framwork}, which parses each trajectory through a context encoder and experience decoder to produce the distilled skill $e = \mathcal{G}(\tau \mid \theta)$. Specifically, after each interaction round~$t$, $\mathcal{J}$ receives the accumulated state~$s_t$ and the agent's latest action~$a_t$, and produces a structured evaluation:
\begin{equation}
    \mathcal{J}(s_t,\, a_t) \;\longrightarrow\; (\ell_t,\; e_t),
    \label{eq:judge}
\end{equation}
where $\ell_t \in \{\texttt{+}, \texttt{-}\}$ is a binary correctness label (positive or negative) and $e_t \in \mathcal{Z}$ is the distilled \emph{logical skill}, which is a natural-language string residing in the abstract skill space~$\mathcal{Z}$ rather than the raw state space~$\mathcal{X}$. The judge is guided by three design principles:

\textbf{Abstraction Protocol.}
Distilled skills must be entity-agnostic: they must not reference specific patient identifiers, concrete API endpoint URLs, or hard-coded numerical values from any particular interaction instance. Because the judge operates exclusively on trajectories collected from a virtual EHR sandbox---where the agent interacts with a simulated FHIR server populated with synthetic patient records rather than real hospital infrastructure---the purpose of this abstraction is not privacy protection but \textit{reusability}: by stripping instance-specific entities, the resulting skills capture transferable reasoning patterns (e.g., parameter mapping logic, age calculation formulas, temporal reasoning strategies) that generalize across tasks and clinical environments. This abstraction map $\alpha: \mathcal{X} \rightarrow \mathcal{Z}$ from the high-dimensional ambient space into a compressed abstract skill space is the mechanism through which SELSM achieves efficient local skill generation: by removing entity-specific details, the same abstraction protocol can be consistently applied at any institution to produce high-quality logical skills from locally collected trajectories, as motivated in Section~\ref{sec:introduction}.

\textbf{Structured Skill Format.}
Each logical skill follows a standardized template: \textit{Task Scenario~$\rightarrow$ Generalized Logic~$\rightarrow$ Canonical Example~$\rightarrow$ Error Avoidance}, covering categories such as Parameter Mapping, Data Extraction, Age Calculation Logic, API Protocol, Temporal Reasoning, and Safety Check.

\textbf{Dual Skill Generation Strategy.}
For positive samples ($\ell_t = \texttt{+}$), the judge generates a \textit{Success Paradigm}~$e_t^{+}$ that captures why the action was correct (e.g., correct mapping logic, appropriate filter conditions, valid calculation methods). For negative samples ($\ell_t = \texttt{-}$), the judge generates a \textit{Correction Axiom}~$e_t^{-}$ that maps error signatures to correct rules, emphasizing constraint conflicts and priority rules. This dual strategy ensures that the skill memory encodes both \emph{what to do} and \emph{what to avoid}.

\subsubsection{Memory Record Construction}
Step-level evaluations sharing the same query~$q$ are aggregated into hierarchical memory records, each pairing a clinical query with its complete set of transition-level tuples:
\begin{equation}
    R_i = \Bigl(q_i,\; \mathcal{T}_i\Bigr), \quad
    \mathcal{T}_i = \bigl\{(s_t^{(i)},\, a_t^{(i)},\, \ell_t^{(i)},\, e_t^{(i)})\bigr\}_{t=1}^{T_i},
    \label{eq:record}
\end{equation}
where $R_i$ denotes the $i$-th record, $q_i$ is its associated query, and $\mathcal{T}_i$ is the transition set containing all step-level tuples that share query~$q_i$.

\subsection{Phase 2: Hierarchical Memory Indexing}
\label{sec:phase2}

Given the collection of memory records $\{R_i\}_{i=1}^{N}$ produced by Phase~1, we construct a two-level embedding index over the skill memory bank using a text embedding model~$\phi$.

\textbf{Task-Level Index.}
For each record~$R_i$, we compute a normalized query embedding:
\begin{equation}
    \mathbf{v}_{q_i} = \hat{\phi}(q_i), \quad \hat{\phi}(\cdot) \triangleq \frac{\phi(\cdot)}{\|\phi(\cdot)\|_2}.
    \label{eq:query_embed}
\end{equation}
Here $\hat{\phi}(\cdot)$ denotes $\ell_2$-normalized embedding, a shorthand we adopt throughout this section. This yields a task-level embedding matrix $\mathbf{V}_Q \in \mathbb{R}^{N \times d}$, where $N$ is the number of unique records and $d$ is the embedding dimension.

\textbf{Transition-Level Index.}
For each transition $(s_t, e_t)$ within record~$R_i$, we compute a state embedding and a skill embedding:
\begin{equation}
    \mathbf{v}_{s_t} = \hat{\phi}(s_t), \qquad
    \mathbf{v}_{e_t} = \hat{\phi}(e_t).
    \label{eq:trans_embed}
\end{equation}
These transition-level embeddings are stored alongside their parent record, enabling fine-grained matching at inference time.

\subsection{Phase 3: Query-Anchored Two-Stage Retrieval and Injection}
\label{sec:phase3}

At inference time, given a new state observation~$s^{*}$, SELSM performs a two-stage hierarchical retrieval, as illustrated in the bottom pipeline of Fig.~\ref{fig:framwork}. This mechanism is designed to resolve the \emph{state polysemy problem} introduced in Section~\ref{sec:introduction}: identical intermediate states (e.g., a successful \texttt{GET} response) can arise across fundamentally different clinical workflows, and purely local state matching would retrieve operationally irrelevant skills.

\subsubsection{Stage 1: Task-Level Filtering}
We first extract the query portion~$q^{*}$ from the current state~$s^{*}$ and score each memory record by cosine similarity in the query embedding space:
\begin{equation}
    \sigma_i^{(q)} = \mathbf{v}_{q^{*}}^{\!\top}\, \mathbf{v}_{q_i}, \quad i = 1, \ldots, N,
    \label{eq:stage1}
\end{equation}
where $\mathbf{v}_{q^{*}} = \hat{\phi}(q^{*})$. The top-$Q$ records ($Q{=}5$ by default) are selected to form the candidate set~$\mathcal{R}^{*}$. This stage anchors the retrieval within the correct \emph{semantic neighborhood}, ensuring that only skills from semantically aligned clinical workflows enter the candidate pool.

\subsubsection{Stage 2: Transition-Level Ranking}
From the candidate records~$\mathcal{R}^{*}$, all transitions are pooled and scored against the current full state:
\begin{equation}
    \sigma_j = \mathbf{v}_{s^{*}}^{\!\top}\, \mathbf{v}_{s_j}, \quad
    \forall\; j \in \bigcup_{R_i \in \mathcal{R}^{*}} \mathcal{T}_i,
    \label{eq:stage2}
\end{equation}
where $\mathbf{v}_{s^{*}} = \hat{\phi}(s^{*})$ and $\sigma_j$ denotes the relevance score of the $j$-th candidate transition. Candidates with $\sigma_j < \theta$ (default $\theta{=}0.9$) are discarded. The remaining transitions are ranked in lexicographic order by $(\sigma_j,\; t_j)$, where $t_j$ is the round index, so that ties in relevance are broken in favor of later-stage, more informative skills. The top-$M$ results ($M{=}1$ by default) are returned.

\subsubsection{Skill Injection}
The retrieved logical skills $\mathcal{E}^{*} = \{e_{j_1}, \ldots, e_{j_M}\}$ are formatted and injected into the agent's prompt alongside the current state~$s^{*}$ before each decision point, serving as structured prior knowledge (Fig.~\ref{fig:framwork}, \textit{Skill Integration \& Decision Making}). This mechanism provides the agent with access to relevant abstract reasoning patterns without modifying the underlying model weights, achieving a form of retrieval-augmented generation (RAG) for agentic clinical decision-making. Crucially, because the retrieved skills reside in the entity-agnostic abstract skill space~$\mathcal{Z}$ rather than the raw state space~$\mathcal{X}$, the same distillation and retrieval methodology can be efficiently replicated at any new institution: domain experts familiar with local clinical workflows generate representative operational queries, the agent collects trajectories within a local simulator, and the resulting institution-specific skills are immediately deployable, all without modifying the underlying model weights.

\subsection{Evaluation Metrics}
\label{sec:metrics}

Let $\mathcal{D} = \{d_1, \ldots, d_N\}$ denote the evaluation set of $N$ clinical tasks. For each task~$d_i$, the agent interacts with the EHR system for $T_i$ turns and produces a terminal response. We define the following primary and derived evaluation metrics. Throughout, $\mathbbm{1}[\cdot]$ denotes the \emph{indicator function}, which returns~$1$ when its predicate is satisfied and~$0$ otherwise.

\subsubsection{Primary Metrics}

\textbf{Task Completion (TC).}
A task is considered \emph{completed} if the agent issues a \texttt{FINISH} command within the allowed interaction budget without triggering an irrecoverable error:
\begin{equation}
    \text{TC} = \frac{1}{N}\sum_{i=1}^{N} \mathbbm{1}\bigl[\mathrm{completed}(d_i)\bigr] \times 100\%.
    \label{eq:tc}
\end{equation}

\textbf{Success Rate (SR).}
A completed task is further deemed \emph{successful} if its terminal answer matches the gold-standard reference:
\begin{equation}
    \text{SR} = \frac{1}{N}\sum_{i=1}^{N} \mathbbm{1}\bigl[\mathrm{success}(d_i)\bigr] \times 100\%.
    \label{eq:sr}
\end{equation}

Since the benchmark partitions tasks into query tasks~$\mathcal{D}_q$ (information retrieval) and action tasks~$\mathcal{D}_a$ (state-modifying operations), with $\mathcal{D} = \mathcal{D}_q \cup \mathcal{D}_a$, we define sub-category success rates:
\begin{equation}
    \text{Q-SR} = \frac{1}{|\mathcal{D}_q|}\sum_{i \in \mathcal{D}_q} \mathbbm{1}\bigl[\mathrm{success}(d_i)\bigr] \times 100\%,
    \label{eq:qsr}
\end{equation}
\begin{equation}
    \text{A-SR} = \frac{1}{|\mathcal{D}_a|}\sum_{i \in \mathcal{D}_a} \mathbbm{1}\bigl[\mathrm{success}(d_i)\bigr] \times 100\%.
    \label{eq:asr}
\end{equation}

\subsubsection{Derived Metrics}

\textbf{One-Shot Correct Rate (OSR).}
Measures the fraction of tasks that are successfully completed within the minimum possible interaction cycle ($T_{\min}$ turns):
\begin{equation}
    \text{OSR} = \frac{1}{N}\sum_{i=1}^{N} \mathbbm{1}\bigl[\mathrm{success}(d_i) \;\wedge\; T_i = T_{\min}\bigr] \times 100\%.
    \label{eq:osr}
\end{equation}

\textbf{Error Robustness (ER).}
Quantifies the agent's resilience to hard failures by measuring the complement of the invalid-action rate~$r_{\mathrm{inv}}$ and the task-limit rate~$r_{\mathrm{lim}}$:
\begin{equation}
    \text{ER} = 1 - r_{\mathrm{inv}} - r_{\mathrm{lim}},
    \label{eq:er}
\end{equation}
where $r_{\mathrm{inv}} = \frac{1}{N}\sum_{i=1}^{N} \mathbbm{1}[\mathrm{invalid}(d_i)]$ and $r_{\mathrm{lim}} = \frac{1}{N}\sum_{i=1}^{N} \mathbbm{1}[\mathrm{timeout}(d_i)]$.

\textbf{Query-Action Balance (QAB).}
Measures the uniformity of performance across task types:
\begin{equation}
    \text{QAB} = 1 - \frac{|\text{Q-SR} - \text{A-SR}|}{\max(\text{Q-SR},\, \text{A-SR})}.
    \label{eq:qab}
\end{equation}

\textbf{Token Cost per Correct Task ($\bar{C}_{\mathrm{tok}}$).}
Measures the average inference cost for successfully completed tasks:
\begin{equation}
    \bar{C}_{\mathrm{tok}} = \frac{1}{|\mathcal{S}|}\sum_{i \in \mathcal{S}} c_i, \quad \mathcal{S} = \bigl\{i : \mathrm{success}(d_i)\bigr\},
    \label{eq:token_cost}
\end{equation}
where $c_i$ denotes the total number of tokens consumed during task~$d_i$.

\section{Results}
\label{sec:results}

\subsection{Experimental Setup}
\label{sec:setup}
To rigorously evaluate the zero-shot deployment capabilities of our framework, we adopt \textbf{MedAgentBench} \cite{jiang2025medagentbench} as our primary evaluation sandbox. Unlike read-only benchmarks such as FHIR-AgentBench \cite{lee2025fhir}, which evaluate only question-answering over FHIR resources, MedAgentBench is a high-fidelity virtual EHR environment covering both read (\texttt{GET}) and write (\texttt{POST}) operations. Importantly, although agents interact with a simulated FHIR server rather than real hospital infrastructure, the benchmark tasks themselves are constructed from real clinical data, ensuring that the evaluation faithfully reflects genuine clinical complexity and operational constraints. Agents must not only retrieve patient information but also execute multi-step clinical actions, including placing medication orders, recording vital signs, and scheduling referrals, under strict JSON schema constraints and dynamic state transitions. This emphasis on knowledge-grounded clinical decision-making, rather than mere data lookup, makes MedAgentBench a particularly rigorous testbed for demonstrating our framework's \textit{Sim-to-Real} transferability and zero-shot execution capability.

\textbf{Implementation Details.}
Three locally deployable foundation models serve as backbone agents: \textbf{Qwen3-30B-A3B}, \textbf{Qwen3-32B}, and \textbf{GLM4-32B}, spanning 30B to 32B parameters and representing the model scale that resource-constrained hospitals can realistically host on commodity hardware.
For the LLM-as-Judge~$\mathcal{J}$ described in Section~\ref{sec:judge}, we employ \textbf{DeepSeek V3.2}, selected for its strong instruction-following fidelity and structured evaluation capabilities, which are essential for producing high-quality step-level correctness labels and entity-agnostic logical skills.
The hierarchical memory index (Section~\ref{sec:phase2}) is built using the \textbf{Qwen3-Embedding-4B} text embedding model (dimensionality $d{=}1024$), which encodes both task-level queries and transition-level states into normalized embedding vectors for cosine-similarity retrieval.
During trajectory collection, each training task is executed $K{=}3$ times with temperature-based sampling to increase trajectory diversity.
For retrieval, we set the task-level candidate count $Q{=}5$, the transition-level return count $M{=}1$, and the state-similarity threshold $\theta{=}0.9$.
The evaluation benchmark comprises 300 clinical tasks spanning both query (information retrieval) and action (state-modifying operations) categories.
All memory-augmented baselines (ExpeL~\cite{zhao2024expel} and A-Mem~\cite{xu2025mem}) are evaluated under identical backbone and trajectory-collection settings to ensure a fair comparison.

\subsection{Overall Performance}
\label{sec:overall}
Table \ref{tab:comparison} presents a side-by-side comparison between the baseline models and our enhanced framework using locally deployable foundation models (ranging from 30B to 32B parameters). The results strongly support our \textit{Accumulating Logical Skills} hypothesis. In the baseline setting, even highly capable models struggle significantly with complex clinical workflows, averaging a Success Rate (SR) of only 41.22\% and an Action Success Rate (A-SR) of 37.77\%. This discrepancy primarily stems from formatting hallucinations and an inability to adhere to unseen local operational rules. By plugging in SELSM—without updating a single model parameter—the performance improves substantially across all metrics. The average SR increases by an absolute 19.67\% (reaching 60.89\%), and the A-SR jumps by nearly 25\%. Notably, for Qwen3-30B-A3B, our framework eliminates task chain breakdowns, driving Task Completed (TC) to an optimal 100.00\%. Furthermore, GLM4-32B, which natively suffers from severe format adherence issues (baseline A-SR: 16.66\%), experiences a 44.00\% absolute improvement when guided by retrieved logical priors. These substantial gains empirically demonstrate that dynamically retrieving entity-agnostic clinical logic acts as an external cognitive scaffold, substantially mitigating the medical data scarcity dilemma.

\begin{table*}[t] 
    \centering
    \begin{adjustbox}{max width=\textwidth}
    \begin{tabular}{
        l           
        cccc        
        c           
        cccc        
    }
        \toprule
        \multirow{2.5}{*}{\textbf{Model}} & 
        \multicolumn{4}{c}{\textbf{Baseline}} & 
        & 
        \multicolumn{4}{c}{\textbf{With SELSM}} \\
        
        \cmidrule(lr){2-5} \cmidrule(lr){7-10}
        
         & 
        \textbf{TC} & \textbf{SR} & \textbf{Q-SR} & \textbf{A-SR} & 
        & 
        \textbf{TC} & \textbf{SR} & \textbf{Q-SR} & \textbf{A-SR} \\
        \midrule

        Qwen3-30B-A3B & 
        85.00 & 48.66 & 52.66 & 44.66 & 
        & 
        \textbf{100.00}\improve{15.00} & 
        \textbf{71.33}\improve{22.67} & 
        \textbf{69.33}\improve{16.67} & 
        \textbf{73.33}\improve{28.67} \\

        Qwen3-32B & 
        89.66 & 51.33 & 50.66 & 52.00 & 
        & 
        \textbf{94.33}\improve{4.67} & 
        \textbf{60.33}\improve{9.00} & 
        \textbf{66.66}\improve{16.00} & 
        \textbf{54.00}\improve{2.00} \\

        GLM4-32B & 
        65.00 & 23.66 & 30.66 & 16.66 & 
        & 
        \textbf{75.66}\improve{10.66} & 
        \textbf{51.00}\improve{27.34} & 
        \textbf{41.33}\improve{10.67} & 
        \textbf{60.66}\improve{44.00} \\

        \midrule 

        \textbf{Average} & 
        79.89 & 41.22 & 44.66 & 37.77 & 
        & 
        \textbf{90.00}\improve{10.11} & 
        \textbf{60.89}\improve{19.67} & 
        \textbf{59.11}\improve{14.45} & 
        \textbf{62.66}\improve{24.89} \\

        \bottomrule
    \end{tabular}
    \end{adjustbox}
    
    \caption{
        \textbf{Side-by-side performance comparison.} 
        Metrics are abbreviated as follows: 
        \textbf{TC}: Task Completed (\%), 
        \textbf{SR}: Success Rate (\%), 
        \textbf{Q-SR}: Query Success Rate (\%), and 
        \textbf{A-SR}: Action Success Rate (\%).
        \textbf{Bold} highlights the superior performance between the two settings. 
        Green arrows (\textcolor{wincolor}{$\uparrow$}) indicate improvement over the baseline, 
        while red arrows (\textcolor{losecolor}{$\downarrow$}) indicate a performance drop.
    }
    \label{tab:comparison}
\end{table*}

\begin{table*}[t] 
    \centering
    \begin{adjustbox}{max width=\textwidth}
    \begin{tabular}{l ccc ccc ccc ccc}
        \toprule
        \multirow{2}{*}{\textbf{Base Model}} & 
        \multicolumn{3}{c}{\textbf{TC}} & 
        \multicolumn{3}{c}{\textbf{SR}} & 
        \multicolumn{3}{c}{\textbf{Q-SR}} & 
        \multicolumn{3}{c}{\textbf{A-SR}} \\
        \cmidrule(lr){2-4} \cmidrule(lr){5-7} \cmidrule(lr){8-10} \cmidrule(lr){11-13}
        & ExpeL & A-Mem & \textbf{Ours} 
        & ExpeL & A-Mem & \textbf{Ours} 
        & ExpeL & A-Mem & \textbf{Ours} 
        & ExpeL & A-Mem & \textbf{Ours} \\
        \midrule
        
        \textbf{Qwen3-30B-A3B} & 99.33 & 100.00 & 100.00 & 58.00 & 47.66 & 71.33 & 54.66 & 30.66 & 69.33 & 61.33 & 64.66 & 73.33 \\
        \textbf{Qwen3-32B}     & 95.00 & 89.33  & 94.33  & 58.00 & 24.00 & 60.33 & 56.66 & 35.33 & 66.66 & 59.33 & 12.66 & 54.00 \\
        \textbf{GLM4-32B}      & 50.00 & 87.00  & 75.66  & 20.66 & 19.66 & 51.00 & 21.33 & 30.00 & 41.33 & 20.00 & 9.33  & 60.66 \\
        \midrule
        \textbf{Average} & 81.44 & \textbf{92.11} & 90.00 & 45.55 & 30.44 & \textbf{60.89} & 44.22 & 32.00 & \textbf{59.11} & 46.89 & 28.88 & \textbf{62.66} \\
        
        \bottomrule
    \end{tabular}
    \end{adjustbox}
    \caption{
        \textbf{Performance comparison across different frameworks.} 
        Each framework (ExpeL\cite{zhao2024expel}, A-Mem\cite{xu2025mem}, and Ours) is evaluated on three base models. 
        The last row shows the average performance across the three models.
    }
    \label{tab:method_comparison}
\end{table*}

\begin{table}[ht]
    \centering
    \begin{adjustbox}{max width=\columnwidth}
    \begin{tabular}{
        l               
        cc              
        c               
        c               
        cccc            
    }
        \toprule
        \multirow{2.5}{*}{\textbf{Method Variant}} &
        \multicolumn{2}{c}{\textbf{Logical Modules}} &
        \multicolumn{1}{c}{\textbf{Retrieval}} &
        &
        \multicolumn{4}{c}{\textbf{Performance}} \\

        \cmidrule(lr){2-3} \cmidrule(lr){4-4} \cmidrule(lr){6-9}

         & \textbf{Logical} & \textbf{Reas} &
           \textbf{Retrieval} &
         & \textbf{TC} & \textbf{SR} & \textbf{Q-SR} & \textbf{A-SR} \\
        \midrule

        \textbf{Ours (Full)} &
        \checkmark & \checkmark & 
        \checkmark & &             
        \textbf{100.00} & \textbf{71.33} & \textbf{69.33} & \textbf{73.33} \\
        \midrule

        w/o Operational Logic &
        -- & \checkmark &         
        \checkmark & &             
        98.66 & 68.33 & 64.66 & 72.00 \\

        w/o Reason &
        \checkmark & -- &         
        \checkmark & &             
        98.00 & 67.33 & 64.00 & 70.66 \\

        w/o 2 Stage Retrieval &
        \checkmark & \checkmark & 
        -- & &                    
        88.33 & 56.66 & 41.33 & 72.00 \\

        \bottomrule
    \end{tabular}
    \end{adjustbox}
    \caption{
        \textbf{Ablation study on the Qwen3-30B-A3B backbone.}
        Effect of removing individual components from the full framework.
        \textbf{Logical}: Operational Logic module, \textbf{Reas}: Reasoning Trace module,
        \textbf{Retrieval}: Two-Stage Retrieval mechanism.
    }
    \label{tab:ablation}
\end{table}

\subsection{Comparison with Memory-Augmented Baselines}
\label{sec:baselines}
To further evaluate the effectiveness of mapping experiences into an entity-agnostic abstract skill space, we compare our framework against recent state-of-the-art (SOTA) agentic memory paradigms, including ExpeL \cite{zhao2024expel} and A-Mem \cite{xu2025mem}. As shown in Table \ref{tab:method_comparison}, existing memory-augmented frameworks yield sub-optimal performance in complex healthcare environments. Strikingly, A-Mem actually degrades the average SR to 30.44\%—which is lower than the vanilla baseline (41.22\%). This severe performance drop empirically corroborates the entity-agnostic abstraction rationale presented in Section~\ref{sec:introduction}. Traditional episodic memory methods store and retrieve raw, noisy textual trajectories from the high-dimensional raw state space $\mathcal{X}$. When faced with novel medical entities, these models suffer from dimensionality explosion and retrieve operationally irrelevant memories that merely share similar surface-level vocabulary. In contrast, our approach limits the search strictly to the cleanly abstracted skill space $\mathcal{Z}$, largely shielding the agent from entity-level semantic hallucinations and establishing a new SOTA average SR of 60.89\%.

\subsection{Ablation Study}
To rigorously isolate the contribution of each framework component and eliminate confounding variables, we conduct our ablation studies exclusively on the strongest performing base model (Qwen3-30B-A3B). Weaker foundation models inherently suffer from severe foundational deficits in long-context comprehension and basic instruction following. Evaluating architectural ablations on such models would entangle the failure of our retrieval/logic modules with the model's intrinsic hallucinations (i.e., the floor effect), thereby obscuring the true scientific impact of the proposed framework. By establishing a clean baseline with the most robust model, we ensure that any observed performance degradation is directly attributable to the removal of our specific components. The ablation results in Table \ref{tab:ablation} detail the impact of removing individual modules. When discarding the internal logical structures (either ``w/o Operational Logic'' or ``w/o Reason''), the Success Rate noticeably drops. This confirms that explicitly decomposing actions into structured operational reasoning helps the LLM better align its internal planning with complex hospital system constraints.

Most importantly, the removal of the two-stage retrieval mechanism (``w/o 2 Stage Retrieval'') results in the most significant performance degradation. The overall SR drops to 56.66\%, and the Query Success Rate (Q-SR) plummets by 28.00\% (from 69.33\% down to 41.33\%). This degradation provides empirical evidence for our \textit{State Polysemy} theory (Section~\ref{sec:phase3}). Without the global query to anchor the semantic neighborhood, purely state-based local retrieval is easily misled by the semantic polysemy of short medical states (e.g., triggering drug retrieval logic merely because of a similar ``search successful'' intermediate output), leading to irreversible failure cascades. Our two-stage mechanism enforces semantic boundaries through hierarchical filtering, ensuring robust sequential execution.

\begin{figure*}[!t]
\centerline{\includegraphics[width=\textwidth]{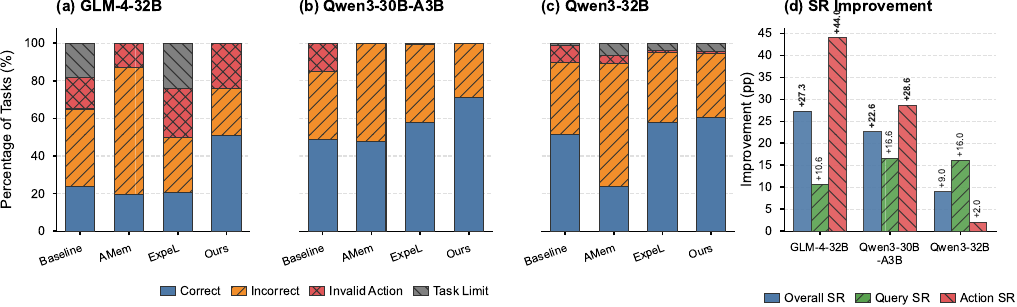}}
\caption{Failure mode distribution and performance improvement analysis. (a)--(c)~Each stacked bar decomposes all 300 tasks into four mutually exclusive outcomes: \textit{Correct} (task completed with the correct answer), \textit{Incorrect} (task completed but with a wrong answer), \textit{Invalid Action} (terminated due to an invalid API call), and \textit{Task Limit} (terminated after exceeding the maximum number of interaction turns) for GLM4-32B, Qwen3-30B-A3B, and Qwen3-32B, respectively. (d)~Absolute improvement of our method over the Baseline in percentage points (pp) across three metrics: Overall Success Rate, Query Success Rate, and Action Success Rate.}
\label{fig:failure_mode}
\end{figure*}

\begin{figure*}[!t]
\centerline{\includegraphics[width=\textwidth]{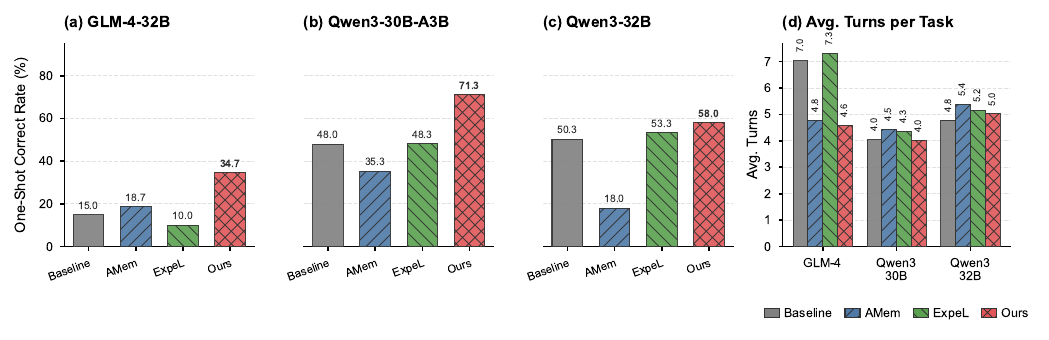}}
\caption{Conversation efficiency analysis. (a)--(c)~One-Shot Correct Rate (OSR; Eq.~\ref{eq:osr}) for each method on GLM4-32B, Qwen3-30B-A3B, and Qwen3-32B, respectively. (d)~Average number of conversation turns per task across all three backbone models.}
\label{fig:conv_efficiency}
\end{figure*}

\subsection{Failure Mode and Robustness Analysis}
\label{sec:failure}

Fig.~\ref{fig:failure_mode}(a)--(c) presents the failure mode distribution for each method across three backbone models.
On Qwen3-30B-A3B, our method achieves a 100\% completion rate with 0\% invalid actions and 0\% task-limit failures, meaning every task terminates normally and produces a parseable answer; 71.3\% of these answers are correct.
In contrast, the Baseline still suffers from 14.7\% invalid actions, while ExpeL and A-Mem retain 0.7\% and 0\% task-limit errors, respectively, yet their incorrect-completion proportions (41.3\% and 52.3\%) remain substantially higher than ours (28.7\%).
On the weaker GLM4-32B backbone, where all methods exhibit higher error rates, our method completely eliminates task-limit failures (0\% vs.\ 18.3\% for the Baseline and 24.0\% for ExpeL), although it shows a comparable invalid-action rate (24.3\%) to ExpeL (26.0\%).
On Qwen3-32B, the four methods show more similar failure profiles; our method achieves the highest success rate (60.3\%) while maintaining low invalid-action (1.3\%) and task-limit (4.3\%) rates.
Overall, panels~(a)--(c) indicate that our method not only improves task success rates but also shifts the failure distribution toward more benign modes by replacing hard failures (invalid actions, timeouts) with softer ones (incorrect but completed answers), particularly on the Qwen3-30B-A3B backbone.

Fig.~\ref{fig:failure_mode}(d) further quantifies the improvement of our method over the Baseline.
On GLM4-32B, our method yields the largest gains, improving the Overall Success Rate by +27.3~pp, Query Success Rate by +10.6~pp, and Action Success Rate by +44.0~pp, the latter representing a near-fourfold increase from 16.7\% to 60.7\%.
On Qwen3-30B-A3B, balanced improvements are observed across all three metrics (+22.6~pp Overall, +16.6~pp Query, +28.6~pp Action).
On Qwen3-32B, the improvement is query-dominant (+16.0~pp Query, +9.0~pp Overall, +2.0~pp Action), suggesting that our memory augmentation provides a stronger boost to information-retrieval tasks on this backbone.
Notably, the improvement magnitude correlates inversely with the backbone's inherent capability: the weaker GLM4-32B benefits the most (+27.3~pp Overall), followed by Qwen3-30B-A3B (+22.6~pp) and Qwen3-32B (+9.0~pp), indicating that our method is especially effective at compensating for the limitations of weaker models.

\begin{figure}[!t]
\centerline{\includegraphics[width=0.9\columnwidth]{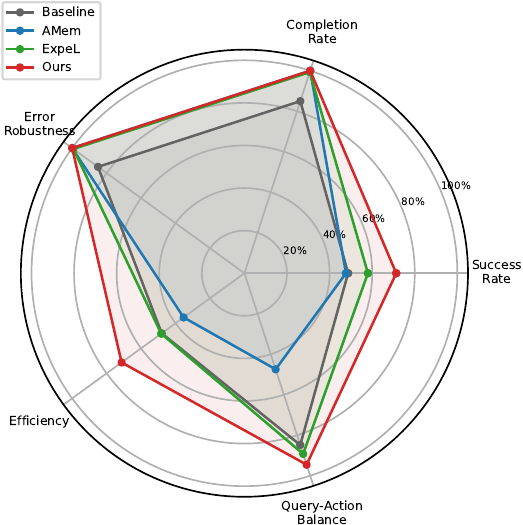}}
\caption{Multi-dimensional comparison of four methods on the Qwen3-30B-A3B backbone. Five normalized dimensions are shown: \textit{Success Rate} ($\text{SR}/100$; Eq.~\ref{eq:sr}), \textit{Completion Rate} ($\text{TC}/100$; Eq.~\ref{eq:tc}), \textit{Error Robustness} (ER; Eq.~\ref{eq:er}), \textit{Efficiency} ($\text{OSR}/100$; Eq.~\ref{eq:osr}), and \textit{Query-Action Balance} (QAB; Eq.~\ref{eq:qab}). A larger enclosed area indicates better overall performance.}
\label{fig:radar}
\end{figure}

Fig.~\ref{fig:radar} provides a multi-dimensional comparison of the four methods on Qwen3-30B-A3B across five normalized metrics.
Our method achieves the highest value on all five dimensions, enclosing the largest radar area.
It attains the maximum possible scores on two dimensions, namely Completion Rate (1.0) and Error Robustness (1.0), reflecting its 100\% completion rate and zero invalid-action and task-limit failures.
On Efficiency (OSR; Eq.~\ref{eq:osr}), our method scores 0.713, substantially outperforming ExpeL (0.483), Baseline (0.480), and A-Mem (0.353).
On Success Rate, our method reaches 0.713, exceeding ExpeL (0.580), Baseline (0.487), and A-Mem (0.477).
On Query-Action Balance, our method scores 0.945, indicating a nearly uniform performance across query tasks (69.3\%) and action tasks (73.3\%).

Among the competing methods, A-Mem exhibits notable weaknesses in both Efficiency (0.353) and Query-Action Balance (0.474): the former reflects its low one-shot correct rate of 35.3\%, while the latter is caused by a large disparity between its Query SR (30.7\%) and Action SR (64.7\%).
The Baseline shows the lowest Completion Rate (0.850) and Error Robustness (0.850), reflecting its 14.7\% invalid-action rate and 0.3\% task-limit rate.
ExpeL achieves a relatively balanced profile with moderate scores across all dimensions but does not match our method on any single metric.
Overall, the radar chart demonstrates that our method delivers the most well-rounded performance, simultaneously maximizing accuracy, reliability, efficiency, and task-type balance.

\subsection{Efficiency Analysis}
\label{sec:efficiency}

Fig.~\ref{fig:conv_efficiency} evaluates the conversation efficiency of all methods using the One-Shot Correct Rate (OSR; Eq.~\ref{eq:osr}).
Our method achieves the highest one-shot correct rate across all three backbone models.
On Qwen3-30B-A3B, our method attains 71.3\%, substantially outperforming the Baseline (48.0\%), ExpeL (48.3\%), and A-Mem (35.3\%).
On GLM4-32B, our method reaches 34.7\%, nearly doubling the best competing method (A-Mem at 18.7\%) and tripling ExpeL (10.0\%).
On Qwen3-32B, our method achieves 58.0\%, ahead of ExpeL (53.3\%), Baseline (50.3\%), and A-Mem (18.0\%).
The average turn count further confirms this advantage: on GLM4-32B, our method requires only 4.6 turns on average compared to 7.0 for the Baseline and 7.3 for ExpeL, while on Qwen3-30B-A3B it achieves the theoretical minimum of 4.0 turns.
These results demonstrate that our memory-augmented approach enables the model to select the correct API call on the first attempt, reducing both the number of interaction rounds and the associated computational cost.

\begin{figure}[!t]
\centerline{\includegraphics[width=\columnwidth]{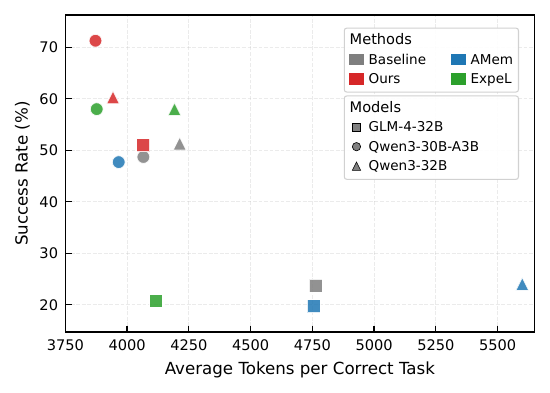}}
\caption{Token efficiency analysis. Each point represents one method--model configuration, with the $x$-axis showing $\bar{C}_{\mathrm{tok}}$ (Eq.~\ref{eq:token_cost}) and the $y$-axis showing SR (Eq.~\ref{eq:sr}). Colors denote methods (Baseline, A-Mem, ExpeL, Ours) and marker shapes denote backbone models (square: GLM4-32B, circle: Qwen3-30B-A3B, triangle: Qwen3-32B). Points closer to the upper-left corner indicate higher accuracy at lower token cost.}
\label{fig:token_efficiency}
\end{figure}

Fig.~\ref{fig:token_efficiency} visualizes the trade-off between token consumption and task accuracy for all method--model configurations. Points in the upper-left region represent the most desirable operating regime: high success rate at low token cost. Across all three backbone models, our method consistently occupies the upper-left region of the plot.
On Qwen3-30B-A3B, our method achieves the highest success rate (71.3\%) while consuming only 3{,}872 tokens per correct task, the lowest among all 12 configurations. ExpeL on the same backbone attains 58.0\% with a comparable token cost (3{,}877), while the Baseline (48.7\%, 4{,}066 tokens) and A-Mem (47.7\%, 3{,}966 tokens) are both lower in accuracy. On GLM4-32B, our method reaches 51.0\% success rate at 4{,}065 tokens per correct task, more than doubling the success rate of the Baseline (23.7\%, 4{,}763 tokens) and ExpeL (20.7\%, 4{,}117 tokens) while using fewer tokens. A-Mem performs the worst on this backbone (19.7\%, 4{,}755 tokens), positioned in the lower-right region of the plot. On Qwen3-32B, our method achieves 60.3\% at 3{,}943 tokens, followed closely by ExpeL (58.0\%, 4{,}192 tokens) and the Baseline (51.3\%, 4{,}213 tokens). A-Mem is a notable outlier on this backbone, consuming the most tokens (5{,}600) while achieving the lowest success rate (24.0\%), making it the least efficient configuration overall. These results demonstrate that our method achieves Pareto-dominant performance: it simultaneously improves task accuracy and reduces per-correct-task token consumption, yielding the most cost-effective inference across all three backbones.

\begin{figure*}[!t]
\centerline{\includegraphics[width=\textwidth]{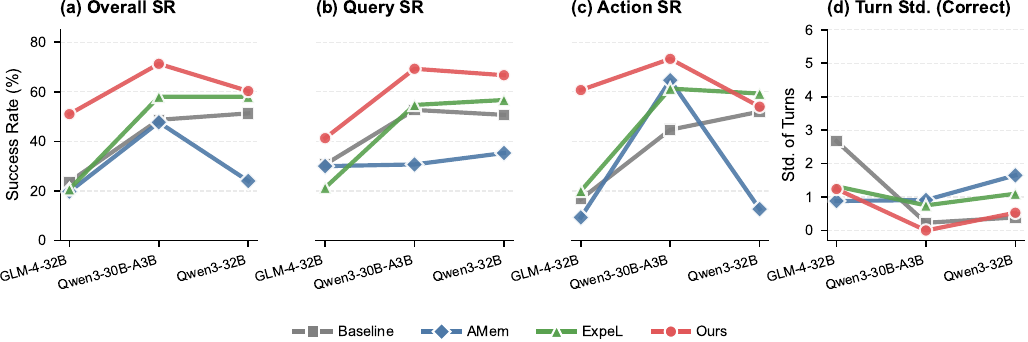}}
\caption{Cross-model generalization analysis. (a)--(c)~Overall, Query, and Action Success Rates of each method across three backbone models. (d)~Standard deviation of conversation turn counts per task (given comparable completion rates, lower values indicate more consistent agent behavior). All methods are evaluated on the same 300-task benchmark.}
\label{fig:cross_model}
\end{figure*}

\subsection{Cross-Model Generalization}
\label{sec:generalization}

Fig.~\ref{fig:cross_model} examines the generalization of all methods across three backbone models of varying capability. Panels~(a)--(c) present the Overall, Query, and Action Success Rates, respectively.
On Overall SR (panel~a), our method achieves the highest accuracy on all three backbones: 51.0\% on GLM4-32B, 71.3\% on Qwen3-30B-A3B, and 60.3\% on Qwen3-32B.
Notably, the improvement over the Baseline is largest on the weakest backbone, GLM4-32B (+27.3~pp), followed by Qwen3-30B-A3B (+22.6~pp) and Qwen3-32B (+9.0~pp), suggesting that our memory augmentation is especially beneficial for less capable models.
A-Mem is particularly unstable across models: it achieves 47.7\% on Qwen3-30B-A3B but drops to 19.7\% on GLM4-32B and 24.0\% on Qwen3-32B, underperforming even the Baseline on both.
ExpeL shows more consistent cross-model behavior, reaching 58.0\% on both Qwen3-30B-A3B and Qwen3-32B, though it achieves only 20.7\% on GLM4-32B.

On Query SR (panel~b), our method leads on all three backbones (41.3\%, 69.3\%, 66.7\%), with A-Mem exhibiting the weakest Query SR across models (30.0\%, 30.7\%, 35.3\%). On Action SR (panel~c), our method achieves the highest scores on GLM4-32B (60.7\%) and Qwen3-30B-A3B (73.3\%). On Qwen3-32B, ExpeL obtains the highest Action SR (59.3\%), slightly above our method (54.0\%) and the Baseline (52.0\%), while A-Mem drops sharply to 12.7\%. Panel~(d) reports the standard deviation of conversation turn counts computed exclusively over correctly completed tasks, measuring how consistently each method solves the tasks it gets right (lower is better). On Qwen3-30B-A3B, our method achieves a standard deviation of 0.00, meaning every correctly answered task is completed in exactly the same number of turns (4 turns), with zero variance. On GLM4-32B, A-Mem shows the lowest standard deviation among correct tasks (0.88), followed by our method (1.24) and ExpeL (1.32), while the Baseline exhibits the highest (2.67).
On Qwen3-32B, the Baseline achieves the lowest standard deviation (0.39), followed by our method (0.53), ExpeL (1.10), and A-Mem (1.65). It is worth noting that methods with fewer correct tasks (e.g., Baseline on GLM4-32B with 71 correct tasks vs.\ Ours with 153) tend to exhibit lower variance because their correct subset is restricted to easier tasks that are uniformly solvable in few turns. In contrast, our method, which solves substantially more tasks including harder ones, naturally incurs slightly higher variance while achieving far superior accuracy.

\section{Discussion}
\label{sec:discussion}

The empirical results of this study validate a central premise: locally deployable foundation models can achieve substantial zero-shot performance gains on complex EHR workflows without environment-specific training data or parameter updates\cite{li2026accurate_rtwh16}. By abstracting simulated clinical interactions into an entity-agnostic skill space of reusable Logical Skills, SELSM equips these models with robust operational capabilities through low-cost, simulation-based skill generation, offering a privacy-preserving and computationally efficient alternative to traditional parameter-updating approaches.

Several cross-cutting patterns in the experimental data illuminate the mechanisms underlying SELSM's effectiveness.
First, improvement magnitude correlates inversely with backbone capability (Table~\ref{tab:comparison}): the weakest backbone benefits most, consistent with a \textit{cognitive scaffolding} interpretation in which logical skills compensate for the reasoning deficits of weaker models---precisely the locally deployable models that resource-constrained hospitals are most likely to adopt.
Second, the ablation study (Table~\ref{tab:ablation}) reveals a clear component hierarchy: removing the two-stage retrieval mechanism causes by far the largest degradation, empirically validating that resolving state polysemy through query-level semantic anchoring is the framework's most critical mechanism.
Third, the failure mode analysis (Fig.~\ref{fig:failure_mode}) shows that SELSM's primary effect is converting hard failures (invalid actions, timeouts) into soft failures (incorrect but completed answers), while achieving Pareto-dominant token efficiency across all backbones (Fig.~\ref{fig:token_efficiency}).
Finally, A-Mem's degradation below the vanilla baseline provides the strongest negative evidence: naive memory augmentation in the raw state space~$\mathcal{X}$ is not merely ineffective but actively harmful in entity-rich medical environments, confirming the necessity of entity-agnostic skill abstraction.

Despite these gains, the results also reveal clear boundary conditions.
On GLM4-32B, our method still exhibits a 24.3\% invalid-action rate, comparable to ExpeL's 26.0\% (Fig.~\ref{fig:failure_mode}a), indicating a \textit{backbone floor effect}: when the base model's fundamental instruction-following and JSON generation capabilities fall below a certain threshold, even high-quality logical skills cannot fully compensate for intrinsic model deficiencies.
Similarly, on Qwen3-32B, ExpeL achieves a higher Action Success Rate (59.3\%) than our method (54.0\%, Table~\ref{tab:method_comparison}), suggesting that for sufficiently capable models on action-type tasks, direct experience replay can occasionally outperform abstract skill matching, likely because write operations benefit more from concrete procedural templates than from generalized logical rules.
These observations collectively delineate the optimal operating regime for SELSM: it provides the greatest advantage when the backbone possesses adequate instruction-following ability but lacks the domain-specific clinical knowledge required for complex multi-step workflows.

We identify five principal limitations that qualify the scope and interpretation of our findings.

\textbf{Single-benchmark evaluation.}
All experiments are conducted on MedAgentBench \cite{jiang2025medagentbench}, which is currently the only sandbox mandating both information retrieval (\texttt{GET}) and state-modifying operations (\texttt{POST}) under realistic API constraints---existing alternatives either evaluate only read-only question-answering \cite{lee2025fhir} or cannot assess multi-step execution \cite{singhal2023large,schmiedmayer2025llmonfhir}.
While this justifies our benchmark choice, reliance on a single evaluation environment inevitably constrains the generalizability of our conclusions to other clinical task types and domains. \textbf{Absence of real cross-institutional validation.} Although the entity-agnostic design of SELSM is theoretically motivated for cross-institutional deployment, all experiments are conducted within a single simulated FHIR environment.
The claimed generalization pathway---whereby each hospital generates institution-specific skills from local simulators---remains a hypothesis awaiting empirical confirmation through multi-site studies with heterogeneous clinical IT infrastructures. \textbf{LLM-as-Judge quality ceiling.}
The LLM-as-Judge mechanism (Section~\ref{sec:judge}) operates on trajectories collected from a virtual EHR sandbox---where agents interact with a simulated FHIR server populated with synthetic patient records---rather than with real hospital infrastructure; its core function is not to handle sensitive clinical information but to distill transferable logical skills from these simulated interactions.
Nevertheless, the quality of extracted skills is fundamentally bounded by the judge model's own reasoning and abstraction capacity. Systematic biases in the judge---such as favoring verbose or syntactically fluent formulations over concise operational rules---may propagate to the skill memory undetected, and the current framework provides no independent calibration mechanism to audit or correct such biases. \textbf{Seed query dependency.} The quality and comprehensiveness of the distilled skill library depend critically on the seed queries used during trajectory collection.
In practice, generating effective and diverse seed queries requires clinicians or clinical informatics specialists with substantial domain expertise, representing a non-trivial human investment at each deployment site. \textbf{Text-only interaction assumption.}
The current framework is designed exclusively for text-based EHR interactions via FHIR APIs.
Extending SELSM to multimodal clinical settings---such as radiology workflows involving DICOM-based medical imaging or intensive care scenarios requiring physiological signal analysis---would necessitate fundamentally new abstraction mechanisms beyond the textual skill framework proposed here.

The entity-agnostic design of SELSM suggests potential applicability beyond the FHIR-based setting validated in this work. Real-world clinical environments involve heterogeneous systems---including PACS/RIS for medical imaging, laboratory information systems using HL7 messaging, and localized hospital information systems with non-standard data dictionaries---each governed by distinct standards and often fragmented by vendor lock-in and historical technical debt \cite{iroju2013interoperability,benson2012principles}.
In principle, the separation of clinical logic from the ambient state space should allow each hospital to generate institution-specific operational experiences within its own simulated environment, then deploy the distilled skills without modifying the underlying model.
Empirically validating this cross-institutional generalization across diverse clinical IT infrastructures, and extending the framework to multimodal clinical workflows, constitute the primary directions for future work.

\section{Conclusion}
\label{sec:conclusion}

In this paper, we introduced the State-Enhanced Logical-Skill Memory (SELSM) framework to address a practical challenge in deploying Medical Agents: how to substantially improve the zero-shot EHR task execution of locally deployable foundation models without environment-specific training data or parameter updates. SELSM decouples clinical reasoning from low-level API semantics by projecting raw interaction trajectories into an entity-agnostic abstract skill space of reusable Logical Skills, and dynamically retrieves the most relevant skills at inference time via a query-anchored two-stage retrieval mechanism. Evaluated on MedAgentBench, SELSM substantially elevates the task completion and success rates of locally deployable foundation models (30B--32B), with Qwen3-30B-A3B achieving 100\% task completion and a 22.67\% absolute gain in success rate. Because the distilled skills are entity-agnostic by design, the framework also offers a promising localization pathway for broader clinical deployment: each hospital could generate operational experiences specific to its own equipment and clinical protocols within a simulated environment, then directly deploy the distilled skills without modifying the underlying model---a direction we plan to validate in future multi-site studies.

\newpage

\section{References}

\bibliographystyle{IEEEtran}
\bibliography{main}

\end{document}